\documentclass{article}

\usepackage[preprint]{neurips_2026}

\usepackage[utf8]{inputenc}
\usepackage[T1]{fontenc}
\usepackage{hyperref}
\usepackage{url}
\usepackage{booktabs}
\usepackage{amsfonts}
\usepackage{amsmath}
\usepackage{graphicx}
\usepackage{xurl}
\usepackage{xcolor}
\usepackage{longtable}
\usepackage{array}
\usepackage{multirow}
\usepackage{framed}

\hypersetup{
    colorlinks=true,
    linkcolor=blue!50!black,
    citecolor=blue!50!black,
    urlcolor=blue!60!black,
    pdftitle={The Metacognitive Profile Atlas},
    pdfauthor={Jon-Paul Cacioli},
}

\title{The Metacognitive Profile Atlas: \\
       Domain-level confidence discrimination in 33 frontier LLMs on MMLU}

\author{%
  Jon-Paul Cacioli \\
  Independent Researcher, Melbourne, Australia \\
  ORCID: 0009-0000-7054-2014 \\
  \url{https://github.com/synthiumjp/metacognitive-profile-atlas}
}

\begin{document}

\maketitle

\begin{abstract}
We release the Metacognitive Profile Atlas, a corpus of answer-plus-verbalized-confidence data from 33 frontier LLMs on 1{,}500 MMLU items stratified into six subject-group bins (46{,}157 scored observations after excluding parser-supplied values). The data allow Type-2 AUROC, the degree to which confidence discriminates correct from incorrect answers, to be decomposed by domain. The reliable finding is a population-level ordering on MMLU. Confidence discriminates best on Applied/Professional items (mean AUROC .750; ranked first in 100\% of bootstrap resamples) and worst on Formal Reasoning and Natural Science (bottom two in 93\% and 92\%; Friedman $\chi^2(5) = 28.26$, $p < .0001$). After Holm correction Applied exceeds Formal, Science, Social and Humanities; no other pair survives, including Applied versus Factual. The ordering holds when imprecise cells are excluded or down-weighted. Per-model six-domain profiles are not reliable (median split-half $r = .16$) and are released for exploration only; the six bins are a pragmatic grouping, not a latent construct. Within-family similarity of profile shape is significant overall, but after multiplicity correction only Anthropic is individually supported. Paired-item analysis separates genuine monitoring gains (Opus 4.1 to 4.5 and 4.7) from error removal (Gemma 3 27B to Gemma 4 31B). Models whose binary-probe confidence failed or could not be certified on a separate battery discriminate well under verbalized confidence, and three Gemma 3 models show the reverse pattern. This version incorporates revisions committed during peer review, corrects descriptions of external validity-screen results, and excludes 980 parser-imputed confidence values for GPT-oss-120B, whose v1 profile was an artefact of them.
\end{abstract}

\begin{framed}
\noindent\textbf{Changes in version 2.} (1) GPT-oss-120B: 980 of its 1{,}500 confidence values were a parser default (50), not model output. Version 1 analysed them and described the result (AUROC .530, confidence SD 21.3) as an anomalous monitoring profile. They are now excluded, along with five such values in four other models and 13 unparsed answers, and all analyses, tables and figures are recomputed, with bootstrap intervals now seeded per cell; on its 518 genuine responses GPT-oss-120B has AUROC .800. (2) Claims about models being ``Invalid'' on a binary-probe battery relied on arXiv:2604.17707/17714v1, which contained a scoring error; corrected classifications (arXiv:2604.17714v2) are used here, and DeepSeek-R1 is not described as inverted. (3) Revisions committed during review are incorporated: claims narrowed to the population-level extremum ordering; precision-weighted, rank-stability and Holm-corrected analyses; paired-item decomposition of generational gains; per-family multiplicity correction; a parsing appendix; and the excluded MMLU subject corrected (\texttt{college\_medicine}, not \texttt{elementary\_mathematics}). Appendix~\ref{app:changes} lists the changed claims.
\end{framed}

\section{Introduction}

The Metacognitive Profile Atlas is a public dataset of (answer, verbalized confidence) pairs from 33 frontier large language models (LLMs) on 1{,}500 stratified items from the Massive Multitask Language Understanding benchmark (MMLU; \citealp{hendrycks2021mmlu}). Items are grouped into six subject-group bins (Applied/Professional, Factual Recall, Formal Reasoning, Humanities/Comprehension, Natural Science, Social/Moral), 250 per bin. Models span eight families, including four Anthropic generations. After excluding parser-supplied values, the corpus holds 46{,}157 scored observations. Data are released under CC-BY-4.0 and code under MIT, with Croissant metadata.

\textbf{Why decompose by domain.} LLM confidence is used for abstention, routing and escalation \citep{wen2025}. Evaluations usually report one aggregate statistic, such as Type-2 AUROC (how well confidence separates correct from incorrect answers) or expected calibration error. An aggregate can hide domains where confidence carries little information. Gemini 2.5 Pro illustrates this. Its aggregate AUROC is .712, yet its confidence discriminates at .889 on Applied items and at chance (.485) on Natural Science items (\S\ref{sec:anomalous}).

\textbf{Background.} Type-2 AUROC is the standard nonparametric measure of metacognitive sensitivity in human research \citep{fleming2014,maniscalco2012}: the probability that a randomly chosen correct answer receives higher confidence than a randomly chosen error. It measures discrimination, not calibration. In LLMs, verbalized and token-based confidence carry some information about correctness \citep{kadavath2022}, but verbalized confidence is often overconfident and concentrated near the top of the scale \citep{xiong2023,yang2024}; \citet{steyvers2025} review the area. Selective prediction, which answers only above a confidence threshold, depends on discrimination \citep{elyaniv2010,geifman2017}. Mechanistic work suggests verbal confidence reflects a distinct internal evaluation \citep{kumaran2026}, separable from calibration directions in the residual stream \citep{miao2026}. Hidden-state signatures of known versus unknown items are also reported \citep{kim2026}. \citet{haznitrama2026} profile capability across many models. The atlas asks a narrower question: whether confidence discrimination differs systematically across content domains, across many models, on one benchmark.

\textbf{Contributions.}
\begin{enumerate}
\item A 33-model dataset of item-level answers and verbalized confidence on 1{,}500 stratified MMLU items, with per-cell error counts, bootstrap intervals, parsing code and an item locator table.
\item A population-level ordering on MMLU: Applied/Professional is the domain where confidence discriminates best, and Formal Reasoning and Natural Science are where it discriminates worst. The ordering survives precision weighting, bootstrap rank analysis and Holm-corrected pairwise tests. The middle domains are not distinguishable.
\item Evidence about what is \emph{not} reliable: per-model six-domain profiles have weak split-half reliability, and the six bins do not form cohesive subject groups. Per-model profiles and family patterns are released as exploratory.
\item A paired-item decomposition showing that equal aggregate gains between model generations can come from opposite mechanisms: genuine monitoring improvement or error removal.
\end{enumerate}

The atlas is a benchmark-stage description on MMLU, not a deployment certification. All claims are specific to MMLU and to one verbalized-confidence prompt.

\section{Methods}

\textbf{Models and items.} Thirty-three LLMs from eight families were evaluated through the Kaggle Benchmarks platform (March--April 2026). The families were Anthropic (Opus 4.1--4.7, Sonnet 4--4.6, Haiku 4.5; $n=8$), DeepSeek (V3.1, V3.2, R1; $n=3$), Google-Gemini (2.0 Flash through 3.1 Pro; $n=7$), Google-Gemma (3 1B through 4 31B; $n=5$), OpenAI (GPT-5.4 nano, mini and full, GPT-oss-20B and 120B; $n=5$), Qwen ($n=4$) and Zhipu (GLM-5). All calls used greedy decoding (temperature 0) and an independent conversation per item. Items are 1{,}500 MMLU test questions drawn deterministically (seed 42), 250 per bin. Fifty-six of 57 MMLU subjects were mapped a priori to the six bins. \texttt{college\_medicine} was left unmapped because it is ambiguous between Natural Science and Applied/Professional. (Version 1 named the excluded subject as \texttt{elementary\_mathematics}, which is in fact mapped to Formal Reasoning.) Table~\ref{tab:domains} summarises the bins. A per-observation (subject, split, row-index) locator table is released.

\begin{table}[!htbp]
\centering
\caption{MMLU-to-domain mapping (partial). Full mapping in the repository.}
\label{tab:domains}
\small
\begin{tabular}{llrr}
\toprule
\textbf{Domain bin} & \textbf{Label} & \textbf{Available items} & \textbf{Sampled} \\
\midrule
Applied/Professional & Applied & 3{,}610 & 250 \\
Factual Recall & Factual & 2{,}998 & 250 \\
Formal Reasoning & Formal & 1{,}910 & 250 \\
Humanities/Comprehension & Human. & 987 & 250 \\
Natural Science & Science & 1{,}532 & 250 \\
Social/Moral & Social & 2{,}832 & 250 \\
\bottomrule
\end{tabular}
\end{table}

\textbf{Elicitation and parsing.} Each item was presented as multiple choice (A--D). The model was asked for its answer and a confidence from 0 to 100 in a fixed format, with no chain-of-thought cue (Appendix~\ref{app:prompt}). Answers and confidences were extracted by a documented rule hierarchy (Appendix~\ref{app:parsing}). Two kinds of record are excluded before analysis. The first is unparsed answers (13 records across six models). The second is confidence values the parser supplied because none could be extracted: the parser substituted 50, and 980 of the 985 such records belong to GPT-oss-120B, which never states 50 itself. The other five are in four models. In every affected model no response genuinely states a confidence of 50, so all its 50s can be identified as parser defaults. These values describe the pipeline, not the model, and including them in version 1 manufactured an apparently anomalous profile for GPT-oss-120B. After exclusion, 46{,}157 observations remain. Sixteen models have fewer than 1{,}500 scored items: twelve because of platform timeouts during the run (the smallest is GLM-5 with 598) and four because of the exclusions. GPT-oss-120B retains 518 items.

\textbf{Analysis.} Type-2 AUROC was computed per model and per model-domain cell from the raw 0--100 confidence. Bootstrap 95\% intervals used 1{,}000 resamples per cell, each cell seeded separately so that its interval depends only on its own data, and per-cell error counts are reported (Table~\ref{tab:s1}). The domain ordering was tested with Friedman's test and Kendall's $W$ over the 33 models. Robustness was checked in four ways: excluding wide-interval or few-error cells, inverse-variance weighting, bootstrap rank stability (500 item resamples within every cell), and Holm-corrected Wilcoxon contrasts for all 15 domain pairs. Ipsative profiles subtract each model's own mean across domains. Family similarity of profile shape compares within-family and between-family correlations of ipsative profiles by permutation (10{,}000 shuffles), with per-family tests Holm-corrected.

\section{Results}

\begin{figure}[!htbp]
\centering
\includegraphics[width=0.78\textwidth]{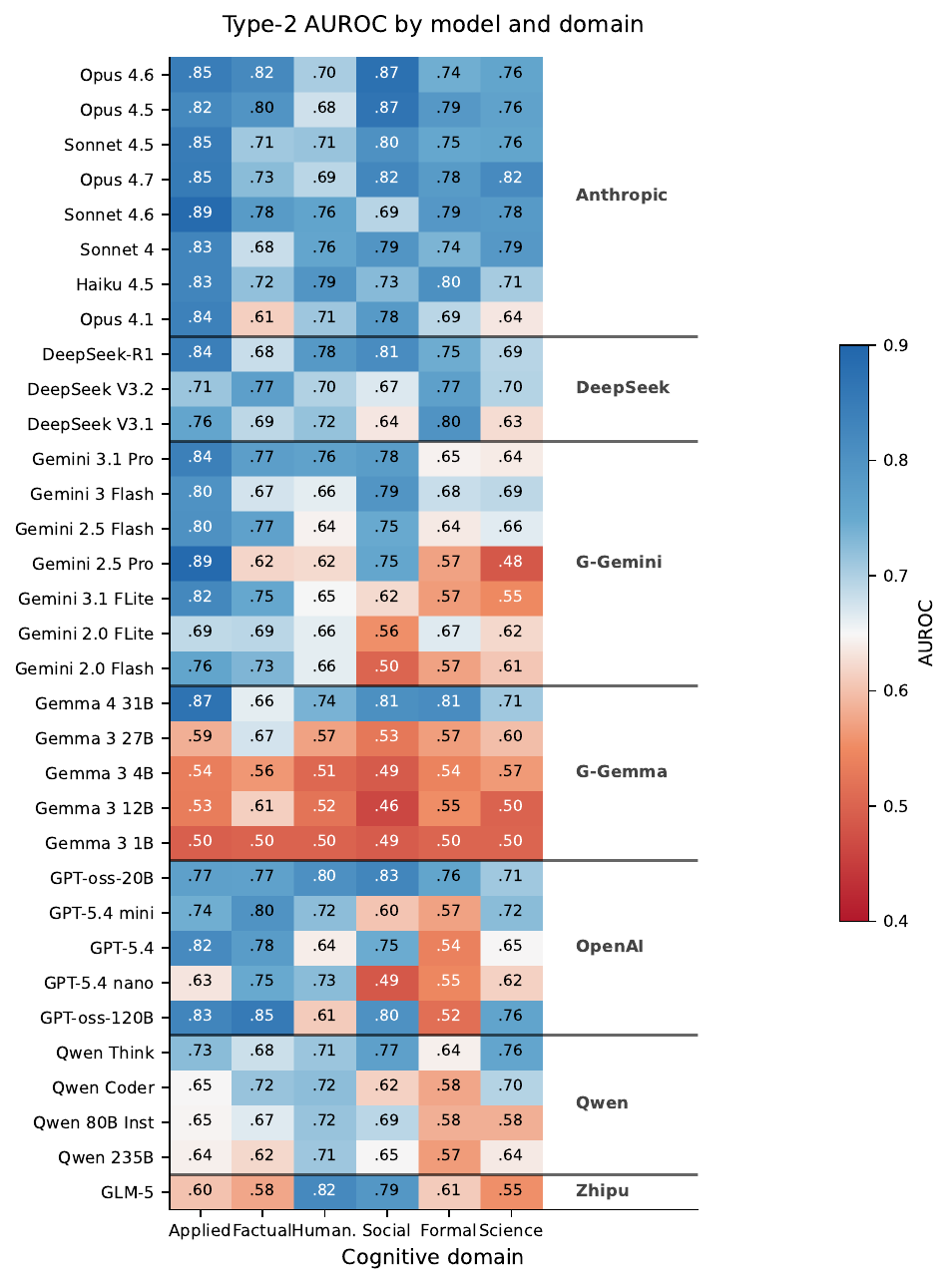}
\caption{Type-2 AUROC by model (rows) and MMLU domain bin (columns). The colour scale spans .40 to .90, so its visual midpoint is about .65, not chance (.50). Family separators are horizontal lines. Full values: Table~\ref{tab:matrix}; intervals and error counts: Table~\ref{tab:s1}.}
\label{fig:heatmap}
\end{figure}

\textbf{Dataset summary.} Accuracy ranges from .388 (Gemma 3 1B) to .951 (Opus 4.6, Gemini 3 Flash). Aggregate AUROC ranges from .498 (Gemma 3 1B) to .806 (Opus 4.6). The 198 model-domain cells have a median of 20 errors (minimum 2; 28 cells have fewer than 10). Their bootstrap intervals have median width .202; 36\% are wider than .25, mostly in high-accuracy models where few errors leave the estimate unstable. Table~\ref{tab:matrix} gives the full matrix and Figure~\ref{fig:heatmap} the heatmap.

\subsection{Domain ordering}
\label{sec:hierarchy}

\begin{figure}[!htbp]
\centering
\includegraphics[width=0.7\textwidth]{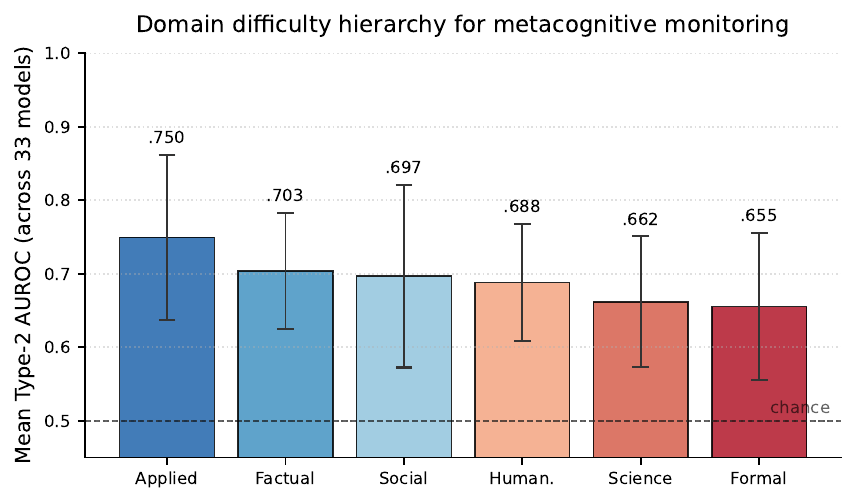}
\caption{Mean Type-2 AUROC per domain across 33 models, with standard deviation bars.}
\label{fig:hierarchy}
\end{figure}

Confidence discriminates best on Applied items and worst on Formal and Science items (Table~\ref{tab:hierarchy}, Figure~\ref{fig:hierarchy}; Friedman $\chi^2(5) = 28.26$, $p < .0001$; Kendall's $W = .171$). Applied ranks in the top two within 21 of 33 models, and Formal or Science occupies a bottom-two rank in 27 of 33. The low concordance means models agree on the extremes, not on the middle.

\begin{table}[!htbp]
\centering
\caption{Domain means across 33 models (Type-2 AUROC).}
\label{tab:hierarchy}
\small
\begin{tabular}{llrrrrr}
\toprule
\textbf{Rank} & \textbf{Domain} & \textbf{Mean AUROC} & \textbf{SD} & \textbf{Mean rank} & \textbf{\#1 in} & \textbf{Last in} \\
\midrule
1 & Applied & .750 & .112 & 2.09 & 15/33 & 0/33 \\
2 & Factual & .703 & .079 & 3.33 & 7/33 & 5/33 \\
3 & Social & .697 & .124 & 3.48 & 4/33 & 9/33 \\
4 & Humanities & .688 & .080 & 3.73 & 4/33 & 4/33 \\
5 & Science & .662 & .089 & 4.21 & 1/33 & 8/33 \\
6 & Formal & .655 & .100 & 4.15 & 2/33 & 7/33 \\
\bottomrule
\end{tabular}
\end{table}

\textbf{Robustness.} The ordering does not depend on imprecise cells (Table~\ref{tab:precision}). Applied stays first and Formal and Science stay in the bottom two under every exclusion rule and under inverse-variance weighting. Under the exclusions, concordance rises ($W$ = .20 to .32), although the Friedman test then rests on fewer complete models. In 500 bootstrap resamples of items within every cell, Applied ranks first in 100\% of resamples, Formal falls in the bottom two in 93\% and Science in 92\%. The gap between Applied and the lowest domain is .098 (95\% interval [.073, .129]). Holm-corrected Wilcoxon tests over the 15 domain pairs show Applied exceeding Formal, Science, Social and Humanities ($p_{\text{Holm}} \le .011$). No other pair survives correction, including Applied versus Factual ($p_{\text{Holm}} = .14$) and Formal versus Science ($p = .87$). Applied is the one reliably distinct domain; Formal and Science form an undifferentiated bottom pair.

\begin{table}[!htbp]
\centering
\caption{Domain ordering under precision-based variants. Friedman tests use only models with all six cells retained.}
\label{tab:precision}
\footnotesize
\setlength{\tabcolsep}{3.5pt}
\begin{tabular}{lrcccccc}
\toprule
\textbf{Variant} & \textbf{Cells} & \textbf{Applied} & \textbf{Formal} & \textbf{Science} & \textbf{Models} & \textbf{Friedman} & \textbf{$W$} \\
\midrule
All cells & 198 & .750 (1st) & .655 (6th) & .662 (5th) & 33 & $\chi^2 = 28.26$, $p < .0001$ & .171 \\
CI width $\le .30$ & 157 & .754 (1st) & .615 (6th) & .659 (5th) & 15 & $\chi^2 = 16.94$, $p = .005$ & .226 \\
CI width $\le .25$ & 127 & .754 (1st) & .586 (6th) & .623 (5th) & 10 & $\chi^2 = 15.83$, $p = .007$ & .317 \\
$\ge 10$ errors & 170 & .754 (1st) & .642 (6th) & .672 (5th) & 18 & $\chi^2 = 18.03$, $p = .003$ & .200 \\
Inverse-variance weighted & 198 & .711 (1st) & .572 (5th) & .534 (6th) & -- & -- & -- \\
\bottomrule
\end{tabular}
\end{table}

\textbf{Accuracy.} The ordering is not explained by domain difficulty in any simple way. Applied has the \emph{lowest} mean accuracy (.828) and Humanities the highest (.922). Across the six domains, mean accuracy and mean AUROC correlate negatively ($\rho = -.31$), and within models the centred correlation is $r = .01$. These checks rule out a monotonic accuracy confound, not every effect of item composition. In high-accuracy cells AUROC rests on few errors, and that sparse-error regime is what the precision analyses address.

\textbf{Reliability: the extremum is reliable, the profile is not.} Aggregate AUROC is reproducible across random item halves (cross-model $r = .868$). The population-level contrast between Applied and the mean of Formal and Science has a split-half correlation across models of .249 (median over 50 splits; IQR .162--.343; Spearman--Brown .40), and its sign is positive in both halves in 50 of 50 splits. Full six-domain profile shapes within a model are much less reliable. Across the same splits the median correlation between halves is .157 (Spearman--Brown .27). Over 100 splits the grand median of per-model medians is .076, and 58\% of models are positive. That last figure is sensitive to individual models because per-model values are spread thinly around it. The claims of this paper therefore rest on the population-level ordering; per-model profiles are released for exploration and should be read against their intervals.

\textbf{Domain bins are not cohesive.} Correlations of per-subject AUROC across models are no higher between subjects in the same bin than between subjects in different bins (mean $r = .173$ in both cases). The bins group MMLU subjects by pragmatic content, not by an empirically cohesive construct. The same limitation shows at the subject level. Applied subjects rank higher on average than Formal and Science subjects (mean rank 19.6 versus 28.8 and 21.1 of 47), but the difference is not significant (Mann--Whitney $p = .18$).

\subsection{Families and generations (exploratory)}

Family-level aggregate AUROC ranges from .573 (Gemma) to .777 (Anthropic) (Table~\ref{tab:families}). Anthropic is the most consistent family (.708--.806 across eight models). In OpenAI, the two open-weight GPT-oss models (.793 and .800) exceed GPT-5.4 (.649), mini (.651) and nano (.582). Version 1 attributed OpenAI's spread to GPT-oss-120B's low AUROC and inferred that training recipe dominates scale; that inference rested on the imputed values and is withdrawn.

\begin{table}[!htbp]
\centering
\caption{Family-level aggregate Type-2 AUROC.}
\label{tab:families}
\small
\begin{tabular}{lrrr}
\toprule
\textbf{Family} & \textbf{\# models} & \textbf{Mean AUROC} & \textbf{Range} \\
\midrule
Anthropic & 8 & .777 & .708--.806 \\
DeepSeek & 3 & .740 & .716--.769 \\
Zhipu & 1 & .697 & -- \\
OpenAI & 5 & .695 & .582--.800 \\
Google-Gemini & 7 & .693 & .623--.765 \\
Qwen & 4 & .669 & .621--.745 \\
Google-Gemma & 5 & .573 & .498--.771 \\
\bottomrule
\end{tabular}
\end{table}

\begin{figure}[!htbp]
\centering
\includegraphics[width=0.78\textwidth]{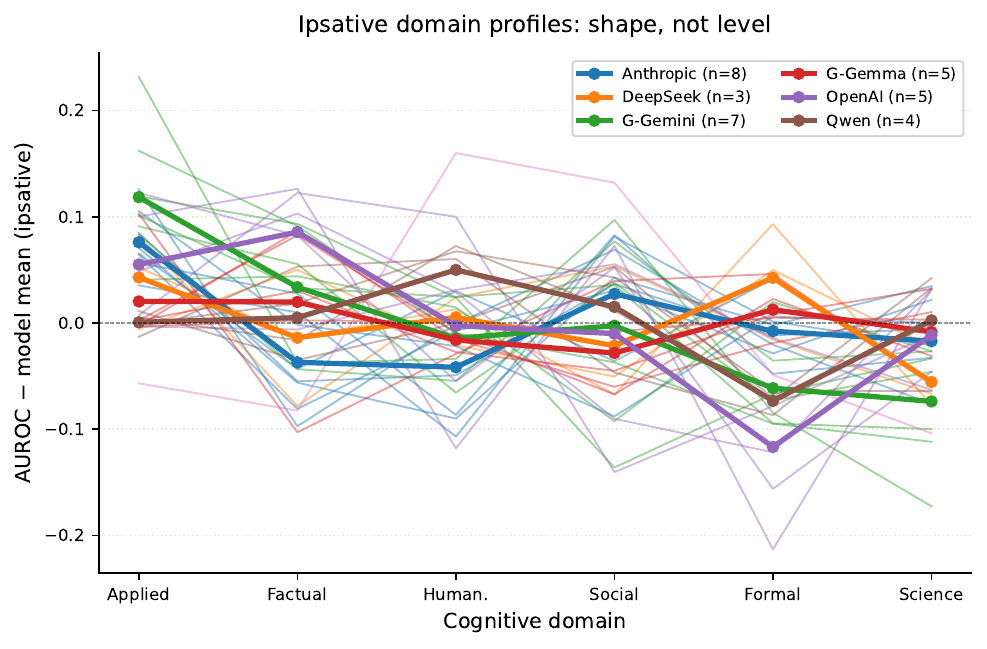}
\caption{Ipsative domain profiles for all 33 models, coloured by family (thin lines: models; bold: family means). Centring on each model's mean isolates profile shape from level. Individual profiles are not reliable (\S\ref{sec:hierarchy}) and are shown for exploration.}
\label{fig:ipsative}
\end{figure}

\textbf{Profile-shape similarity within families.} Ipsative profiles correlate more within families than between them: the mean within-family $r$ is .406 against .091 between, a difference of .316 (permutation $p = .0001$). Per family, only Anthropic is individually supported after Holm correction (Table~\ref{tab:family}). Gemini and Qwen are nominally significant and are reported as exploratory patterns. Given the weak reliability of individual profiles, these family patterns should be treated with the same caution. Figure~\ref{fig:ipsative} shows the profiles.

\begin{table}[!htbp]
\centering
\caption{Per-family permutation tests of profile-shape similarity (10{,}000 permutations; statistic = within-family minus between-family mean correlation), Holm-corrected across six families. Zhipu ($n = 1$) is not testable.}
\label{tab:family}
\small
\begin{tabular}{lrrrrrr}
\toprule
\textbf{Family} & \textbf{$n$} & \textbf{Within $r$} & \textbf{Between $r$} & \textbf{Statistic} & \textbf{$p$} & \textbf{$p_{\text{Holm}}$} \\
\midrule
Anthropic & 8 & .446 & .085 & .360 & .003 & .019 \\
Qwen & 4 & .482 & .025 & .458 & .018 & .092 \\
Google-Gemini & 7 & .511 & .232 & .279 & .019 & .092 \\
Google-Gemma & 5 & .210 & $-$.046 & .256 & .066 & .197 \\
OpenAI & 5 & .310 & .156 & .154 & .155 & .310 \\
DeepSeek & 3 & .125 & .065 & .060 & .370 & .370 \\
\bottomrule
\end{tabular}
\end{table}

\begin{figure}[!htbp]
\centering
\includegraphics[width=\textwidth]{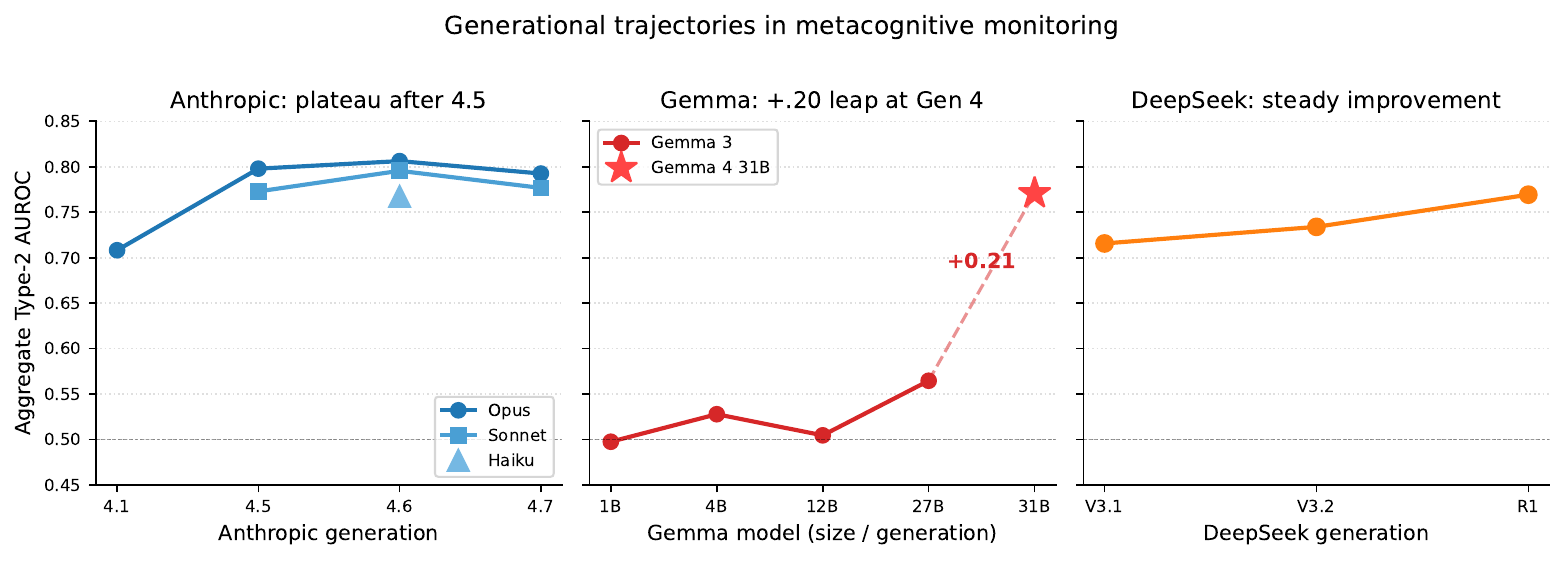}
\caption{Aggregate Type-2 AUROC across generations for Anthropic, Gemma and DeepSeek. Table~\ref{tab:generational} decomposes the changes.}
\label{fig:generational}
\end{figure}

\textbf{Generational change: monitoring gain versus error removal.} Aggregate AUROC rises across several model generations (Figure~\ref{fig:generational}). The Opus line goes from .708 (4.1) to .798 (4.5) and then stays flat (.806, .792) despite continued accuracy gains. Gemma 3 27B (.565) to Gemma 4 31B (.771) is the largest single step. A higher AUROC can arise in two ways: the newer model doubts errors it still makes (better monitoring), or it stops making the confidently wrong answers that depressed the older model's AUROC (error removal). Table~\ref{tab:generational} separates the two on the items the earlier model answered wrongly with confidence of at least 70.

\begin{table}[!htbp]
\centering
\caption{Paired-item decomposition on items the earlier model answered wrongly with confidence $\ge 70$. ``Fixed'': share answered correctly by the later model. $\Delta$conf: mean change in confidence on items still wrong (Wilcoxon signed-rank).}
\label{tab:generational}
\footnotesize
\setlength{\tabcolsep}{4pt}
\begin{tabular}{lrrrrl}
\toprule
\textbf{Pair} & \textbf{Items} & \textbf{Fixed} & \textbf{$\Delta$conf (still wrong)} & \textbf{$p$} & \textbf{Reading} \\
\midrule
Gemma 3 27B $\to$ Gemma 4 31B & 321 & 82.6\% & $-$2.7 & .052 & Error removal \\
Opus 4.1 $\to$ Opus 4.5 & 80 & 35.0\% & $-$7.9 & $3.1\times10^{-7}$ & Monitoring gain \\
Opus 4.1 $\to$ Opus 4.7 & 80 & 45.0\% & $-$6.6 & $3.3\times10^{-5}$ & Monitoring gain \\
Opus 4.5 $\to$ Opus 4.7 & 66 & 37.9\% & $+$0.2 & .59 & Neither \\
DeepSeek V3.1 $\to$ V3.2 & 160 & 37.5\% & $-$2.0 & .002 & Weak monitoring gain \\
DeepSeek V3.2 $\to$ R1 & 135 & 47.4\% & $-$1.8 & .28 & Error removal \\
\bottomrule
\end{tabular}
\end{table}

The Gemma step is mainly error removal. Gemma 4 answers 82.6\% of those items correctly, its confidence on them rises (94.8 to 97.2), and on its 32 newly wrong items it reports a mean confidence of 90.8. The Opus steps show the signature of better monitoring. Opus 4.5 lowers its confidence on the errors that persist (89.1 to 81.2) and even on the errors it fixes (85.7 to 77.9), and Opus 4.7's 31 newly wrong items carry a mean confidence of only 71.5. Similar aggregate gains can therefore come from opposite mechanisms, which is the main reason for releasing paired item-level data. Gemma 3 1B's domain intervals are the narrowest in the atlas (all widths at most .06), so its near-chance aggregate (.498) is a precise estimate of absent discrimination, not noise.

\subsection{Binary-probe validity screening and verbalized confidence}
\label{sec:probe}

Twenty of the 33 models were also tested on a separate 524-item battery that asks after each answer whether to KEEP or WITHDRAW it \citep{cacioli2026c}. A portable validity screen \citep{cacioli2026e} classifies a binary confidence signal from the correctness-by-confidence table. It is Invalid when a response set is established: confidence kept on at least 95\% of errors (L), or withdrawn on at least half of correct answers (Fp), with interval guards. It is Indeterminate when such a pattern cannot be certified, often because there are too few errors, or when confidence does not measurably track correctness. It is Valid otherwise. The corrected classification (arXiv:2604.17714v2) uses the battery's five keyed tracks. It finds DeepSeek-R1 Invalid: it withdraws 95\% of its correct answers, and with only six errors whether its withdrawals track correctness cannot be assessed. Gemini 3.1 Pro and Qwen3 Next 80B Thinking are Indeterminate: they keep confidence on 96\% of their errors but make 24 and 27 errors, too few to certify blanket confidence. Gemini 3 Flash and GPT-5.4 nano are also Indeterminate, with no detectable item sensitivity. Version 1 of this paper described R1 as showing catastrophic inversion; that description came from a scoring error in the battery analysis and is withdrawn.

\begin{figure}[!htbp]
\centering
\includegraphics[width=0.72\textwidth]{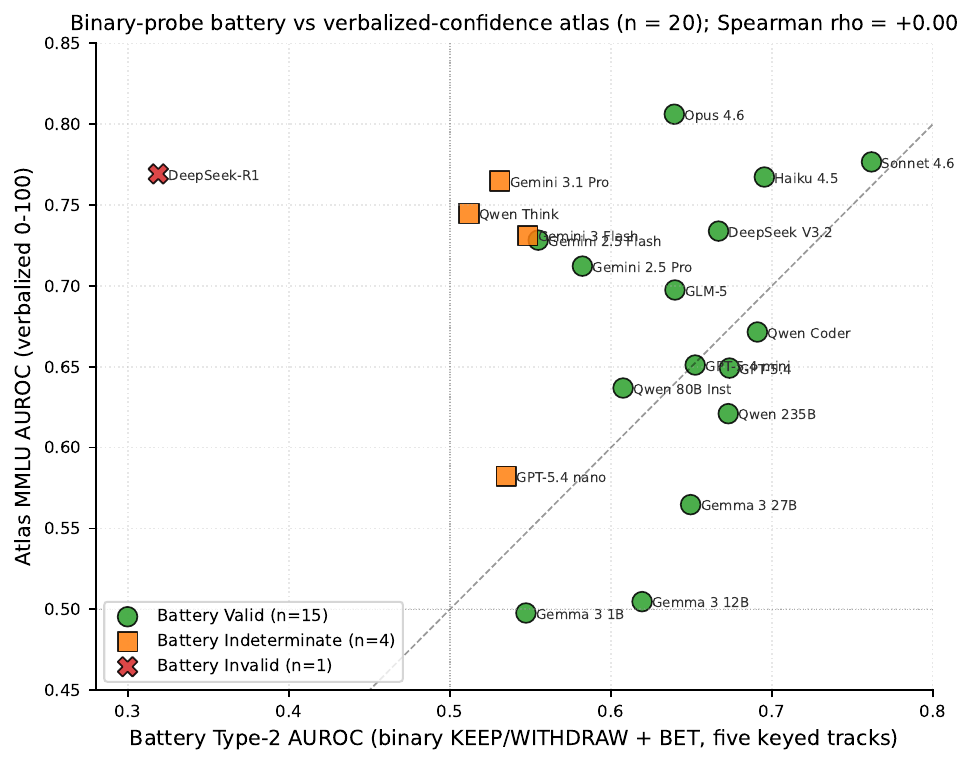}
\caption{Battery Type-2 AUROC (binary KEEP/WITHDRAW + BET probes, five keyed tracks; \citealp{cacioli2026e}) against atlas aggregate AUROC (verbalized 0--100 confidence on MMLU) for the 20 models in both. Markers show the corrected battery tier. The dashed line is identity.}
\label{fig:cross}
\end{figure}

Under verbalized confidence the picture differs in both directions (Figure~\ref{fig:cross}). R1, Gemini 3.1 Pro and Qwen Thinking discriminate well on MMLU (aggregate AUROC .769, .765 and .745; domain ranges .682--.843, .640--.841 and .644--.768). Gemma 3 1B, 12B and 27B are Valid on the binary battery, yet their verbalized confidence barely discriminates on MMLU (.498, .505 and .565). Across the 20 models, battery AUROC does not predict atlas AUROC (Spearman $\rho = .00$). Screened on MMLU at a cut point of 90 (confidence $\ge 90$ counts as high), the four Gemma 3 models are Invalid, Gemini 2.5 Pro is Indeterminate, and the other 28 are Valid. A model's confidence quality is therefore specific to the probe format and task under which it is measured. A binary-probe result says little about verbalized confidence on another benchmark, and the reverse.

\subsection{Individual profiles}
\label{sec:anomalous}

These descriptions are exploratory (\S\ref{sec:hierarchy}).

\textbf{Gemini 2.5 Pro} has the widest within-model spread: Applied .889 [.794, .960], Science .485 [.470, .497], Formal .573 [.478, .778]. The Applied lower bound exceeds the Science upper bound, so the difference is not sampling noise. The aggregate (.712) gives no sign that its Science-domain confidence is at chance. This run is partial (981 items).

\textbf{GLM-5} is the only model whose Humanities AUROC (.818 [.706, .972]) is well above its Applied AUROC (.602). Its run is partial (598 items), and its Humanities estimate rests on three errors. A complete rerun three months later shows what can and cannot be concluded. On the 598 items common to both runs, correctness agreed on 98.5\% of items, item-level confidence correlated $r = .71$, and the six-domain ordering correlated $\rho = .77$ ($p = .072$, six domains). Humanities again exceeded Applied (.792 versus .639), and the aggregate AUROC rose from .697 to .759. The rerun's errors per domain numbered between 2 and 12, so its domain estimates are imprecise as well (Appendix~\ref{app:glm5}).

\textbf{GPT-oss-120B.} Version 1 reported this model as having the highest confidence variability in the atlas with near-chance discrimination (SD 21.3, AUROC .530), and offered it as a distinct failure mode. That profile was produced by 980 parser-imputed values of 50. On its 518 genuine responses its aggregate AUROC is .800 and its domain values range from .515 (Formal) to .855 (Factual). Its imputation rate also differs by domain (59\% to 74\%; $\chi^2(5) = 18.67$, $p = .002$), so the retained items are not a random sample. With 65 to 102 items per domain and wide intervals (Formal: [.128, .918]), its profile should not be interpreted.

\subsection{Validation summary}
\label{sec:coherence}

Aggregate AUROC rankings are reproducible on half the items ($r = .868$; Appendix Figure~\ref{fig:splithalf}). The population-level extremum contrast is reproducible in sign but modest in reliability. Per-model profiles are not reliable, and the domain bins are not cohesive (\S\ref{sec:hierarchy}). The paper's claims are limited accordingly.

\section{Discussion}

\textbf{What the atlas shows.} On MMLU, verbalized confidence across 33 frontier models discriminates correct from incorrect answers best on applied professional knowledge and worst on formal reasoning and natural science. The pattern holds at the population level under precision weighting, bootstrap rank analysis and multiplicity-corrected contrasts. One plausible account: Applied items (law, medicine, accounting) tend to have answers anchored in specific retrieved knowledge, so whether a model knows the answer may leave a clearer internal trace. Formal items admit plausible intermediate steps that end at wrong answers. This fits the hypothesis of \citet{kim2026} that well-learned factual material produces more separable internal states. Nothing here examines hidden states, so the account is a hypothesis, not a mechanism.

\textbf{What it does not show.} It does not show that any individual model has a stable domain profile: per-model profiles have weak split-half reliability. It does not show that the six bins are cognitive constructs: subjects within a bin are no more alike than subjects across bins. It does not show that the ordering holds beyond MMLU or beyond this prompt; replication on a benchmark with non-overlapping items (for example GPQA) is the necessary next test. Type-2 AUROC measures discrimination, not calibration.

\textbf{Implications for confidence-based abstention.} If a deployment applies one confidence threshold across content types, the population-level result predicts that the threshold will be less informative on formal and scientific content than on applied professional content. This argues for checking discrimination separately on the content a system will see, rather than relying on a single aggregate. It does not support looking up a particular model's domain profile in this atlas and treating it as reliable. In flagship models, adjacent-domain differences are not interpretable at current item counts: GPT-5.4's Formal AUROC is .539 [.452, .631], for example. Those comparisons need many more errors per cell.

\textbf{Probe format.} Classifications from a binary KEEP/WITHDRAW probe did not predict discrimination under verbalized confidence, in either direction (\S\ref{sec:probe}). Statements about a model's confidence should name the probe and task: ``no detectable signal under probe X on task Y'', not ``no signal''.

\textbf{Parsing is part of the measurement.} The largest correction in this version came from parser-supplied values being treated as model output. Releasing raw outputs, extraction code and per-model fallback counts made the error findable. We recommend that verbalized-confidence datasets report imputation and fallback rates per model, and exclude imputed values before analysis.

\textbf{Limitations.} The results cover MMLU only and one verbalized-confidence prompt, with greedy decoding. The domain mapping is a priori and not empirically cohesive. Per-cell estimates are imprecise for high-accuracy models, with 36\% of intervals wider than .25 and 28 cells with fewer than 10 errors. Sixteen runs are partial (twelve from platform timeouts, four from exclusions), and GPT-oss-120B's retained items are a non-random subset. Answers and confidences were extracted by rules, with fallbacks that can occasionally capture an unintended letter or number (Appendix~\ref{app:parsing}). No causal mechanism is tested.

\section{Conclusion}

Aggregate confidence metrics hide a reliable population-level pattern on MMLU: verbalized confidence in frontier models discriminates best on applied professional knowledge and worst on formal reasoning and natural science. Per-model profiles are too unreliable to interpret individually at this item count. The atlas releases the item-level data, error counts, parsing code and locators needed to test these patterns on other benchmarks and probes.

\section*{Open science}

Data (item-level CSVs, error counts, bootstrap intervals, MMLU locators, parsing fallback counts), extraction and analysis code, and a single script that regenerates every number and figure are at \url{https://github.com/synthiumjp/metacognitive-profile-atlas} (MIT, code) and \url{https://huggingface.co/datasets/synthiumjp/metacognitive-profile-atlas} (CC-BY-4.0, data; Croissant metadata).

\section*{Use of AI}

Claude (Anthropic) assisted with benchmark design, analysis code and manuscript preparation. Analysis code was checked against \texttt{scikit-learn} reference implementations and by re-running the full pipeline from raw data; the corrections in this version were found by such re-analysis. All scientific decisions, the domain mapping and the interpretations are the author's.

\bibliographystyle{plainnat}
\bibliography{references}

\begin{thebibliography}{17}
\providecommand{\natexlab}[1]{#1}
\providecommand{\url}[1]{\texttt{#1}}
\expandafter\ifx\csname urlstyle\endcsname\relax
  \providecommand{\doi}[1]{doi: #1}\else
  \providecommand{\doi}{doi: \begingroup \urlstyle{rm}\Url}\fi

\bibitem[Cacioli(2026{\natexlab{a}})]{cacioli2026c}
Jon-Paul Cacioli.
\newblock The {M}etacognitive {M}onitoring {B}attery: {A} cross-domain
  benchmark for {LLM} self-monitoring.
\newblock \emph{arXiv preprint arXiv:2604.15702}, 2026{\natexlab{a}}.

\bibitem[Cacioli(2026{\natexlab{b}})]{cacioli2026e}
Jon-Paul Cacioli.
\newblock Screen before you interpret: {A} portable validity screen for
  benchmark-based {LLM} confidence signals.
\newblock \emph{arXiv preprint arXiv:2604.17714v2}, 2026{\natexlab{b}}.

\bibitem[El-Yaniv and Wiener(2010)]{elyaniv2010}
Ran El-Yaniv and Yair Wiener.
\newblock On the foundations of noise-free selective classification.
\newblock \emph{Journal of Machine Learning Research}, 11:\penalty0 1605--1641,
  2010.

\bibitem[Fleming and Lau(2014)]{fleming2014}
Stephen~M. Fleming and Hakwan~C. Lau.
\newblock How to measure metacognition.
\newblock \emph{Frontiers in Human Neuroscience}, 8:\penalty0 443, 2014.

\bibitem[Geifman and El-Yaniv(2017)]{geifman2017}
Yonatan Geifman and Ran El-Yaniv.
\newblock Selective classification for deep neural networks.
\newblock In \emph{Advances in Neural Information Processing Systems},
  volume~30, 2017.

\bibitem[Haznitrama et~al.(2026)Haznitrama, Ardi, and Oh]{haznitrama2026}
Faiz~Ghifari Haznitrama, Faeyza~Rishad Ardi, and Alice Oh.
\newblock A neuropsychologically grounded evaluation of {LLM} cognitive
  abilities.
\newblock \emph{arXiv preprint arXiv:2603.02540}, 2026.

\bibitem[Hendrycks et~al.(2021)Hendrycks, Burns, Basart, Zou, Mazeika, Song,
  and Steinhardt]{hendrycks2021mmlu}
Dan Hendrycks, Collin Burns, Steven Basart, Andy Zou, Mantas Mazeika, Dawn
  Song, and Jacob Steinhardt.
\newblock Measuring massive multitask language understanding.
\newblock \emph{International Conference on Learning Representations}, 2021.

\bibitem[Kadavath et~al.(2022)Kadavath, Conerly, Askell, Henighan, Drain,
  Perez, Schiefer, Hatfield-Dodds, DasSarma, Tran-Johnson,
  et~al.]{kadavath2022}
Saurav Kadavath, Tom Conerly, Amanda Askell, Tom Henighan, Dawn Drain, Ethan
  Perez, Nicholas Schiefer, Zac Hatfield-Dodds, Nova DasSarma, Eli
  Tran-Johnson, et~al.
\newblock Language models (mostly) know what they know.
\newblock \emph{arXiv preprint arXiv:2207.05221}, 2022.

\bibitem[Kim(2026)]{kim2026}
Jaehwan Kim.
\newblock Knowing before speaking: {I}n-computation metacognition precedes
  verbal confidence in large language models.
\newblock Preprints.org, posted 3 April 2026, 2026.
\newblock Not peer-reviewed.

\bibitem[Kumaran et~al.(2026)Kumaran, Conmy, Barbero, Osindero, Patraucean, and
  Veli{\v{c}}kovi{\'c}]{kumaran2026}
Dharshan Kumaran, Arthur Conmy, Federico Barbero, Simon Osindero, Viorica
  Patraucean, and Petar Veli{\v{c}}kovi{\'c}.
\newblock How do {LLMs} compute verbal confidence?
\newblock \emph{arXiv preprint arXiv:2603.17839}, 2026.

\bibitem[Larrabee(2012)]{larrabee2012}
Glenn~J. Larrabee.
\newblock \emph{Forensic Neuropsychology: {A} Scientific Approach}.
\newblock Oxford University Press, 2nd edition, 2012.

\bibitem[Maniscalco and Lau(2012)]{maniscalco2012}
Brian Maniscalco and Hakwan Lau.
\newblock A signal detection theoretic approach for estimating metacognitive
  sensitivity from confidence ratings.
\newblock \emph{Consciousness and Cognition}, 21\penalty0 (1):\penalty0
  422--430, 2012.

\bibitem[Miao and Ungar(2026)]{miao2026}
Miranda~Muqing Miao and Lyle Ungar.
\newblock Closing the confidence-faithfulness gap in large language models.
\newblock \emph{arXiv preprint arXiv:2603.25052}, 2026.

\bibitem[Steyvers and Peters(2025)]{steyvers2025}
Mark Steyvers and Megan A.~K. Peters.
\newblock Metacognition and uncertainty communication in humans and large
  language models.
\newblock \emph{Current Directions in Psychological Science}, 2025.

\bibitem[Wen et~al.(2025)Wen, Bansal, Semnani, and Lam]{wen2025}
Bingbing Wen, Hritik Bansal, Sina~J. Semnani, and Monica~S. Lam.
\newblock Know your limits: {A} survey of abstention in large language models.
\newblock \emph{Transactions of the Association for Computational Linguistics},
  13:\penalty0 529--556, 2025.

\bibitem[Xiong et~al.(2023)Xiong, Hu, Lu, Li, Fu, He, and Hooi]{xiong2023}
Miao Xiong, Zhiyuan Hu, Xinyang Lu, Yifei Li, Jie Fu, Junxian He, and Bryan
  Hooi.
\newblock Can {LLMs} express their uncertainty? {A}n empirical evaluation of
  confidence elicitation in {LLMs}.
\newblock \emph{arXiv preprint arXiv:2306.13063}, 2023.

\bibitem[Yang et~al.(2024)Yang, Tsai, and Yamada]{yang2024}
Daniel Yang, Yao-Hung~Hubert Tsai, and Makoto Yamada.
\newblock On verbalized confidence scores for {LLMs}.
\newblock \emph{arXiv preprint arXiv:2412.14737}, 2024.

\end{thebibliography}

\appendix
\renewcommand{\thetable}{A\arabic{table}}
\renewcommand{\thefigure}{A\arabic{figure}}
\setcounter{table}{0}
\setcounter{figure}{0}
\makeatletter
\renewcommand{\theHtable}{A\arabic{table}}
\renewcommand{\theHfigure}{A\arabic{figure}}
\makeatother

\section{Changes from version 1}
\label{app:changes}

Table~\ref{tab:changes} lists every claim in version 1 that this version withdraws or materially changes. Version 1 remains available on arXiv for the record.

\begin{table}[!htbp]
\centering
\caption{Claims changed between version 1 and version 2.}
\label{tab:changes}
\small
\begin{tabular}{p{0.44\textwidth}p{0.50\textwidth}}
\toprule
\textbf{Version 1} & \textbf{Version 2} \\
\midrule
GPT-oss-120B shows the highest confidence variability (SD 21.3) with near-chance discrimination (AUROC .530) and below-chance Science monitoring, a distinct failure mode. & Withdrawn. 980 of its 1{,}500 confidence values were the parser default of 50. On its 518 genuine responses AUROC is .800. The retained items are a non-random subset, and the domain profile is not interpreted (\S\ref{sec:anomalous}, Appendix~\ref{app:gptoss}). \\
\addlinespace
Within OpenAI, GPT-oss-20B outperforms the six-times-larger GPT-oss-120B, so training recipe dominates scale. & Withdrawn; it rested on the imputed values. \\
\addlinespace
DeepSeek-R1, Gemini 3.1 Pro and Qwen Think are battery-Invalid; R1 shows catastrophic inversion on the binary probe. & Corrected from arXiv:2604.17714v2. R1 is Invalid because it withdraws 95\% of its correct answers; it is not inverted. Gemini 3.1 Pro and Qwen Think are Indeterminate (blanket confidence suspected but not certifiable on 24 and 27 errors). \\
\addlinespace
Tier-level assignments are preserved across probe formats in every battery-Valid case. & Withdrawn. Battery AUROC does not predict atlas AUROC ($\rho = .00$), and three battery-Valid Gemma 3 models barely discriminate on MMLU (\S\ref{sec:probe}). \\
\addlinespace
Introduction: every Valid model in a 20-model battery sample had a domain AUROC below .55 \citep{cacioli2026c}. & Removed; it relied on the battery scoring corrected in arXiv:2604.17714v2. \\
\addlinespace
The Gemma 3 $\to$ Gemma 4 step shows that a single generational step can recover an entire class of failure modes. & Reframed. The paired-item decomposition shows that step is mainly error removal, not improved monitoring (Table~\ref{tab:generational}). \\
\addlinespace
Profile-level split-half: grand median $r = .184$, 61\% of models positive; aggregate split-half $r = .893$. & Recomputed: grand median $r = .076$, 58\% positive; aggregate $r = .868$. Claims rest on the population-level extremum ordering. \\
\addlinespace
Friedman $\chi^2(5) = 27.04$, $W = .164$; 47{,}151 observations. & $\chi^2(5) = 28.26$, $W = .171$; 46{,}157 observations. The ordering is unchanged. \\
\addlinespace
The excluded MMLU subject is \texttt{elementary\_mathematics}. & The excluded subject is \texttt{college\_medicine}; \texttt{elementary\_mathematics} is mapped to Formal Reasoning. The analysis was always run with the correct mapping; only the text was wrong. \\
\addlinespace
Cross-study comparison with MMLU AUROCs computed from median-binarised confidence. & Removed. Median binarisation is superseded in the validity screen \citep{cacioli2026e}, whose verdicts depend on tie handling. \\
\addlinespace
Figure 1 colour scale described as diverging around chance (.50). & The scale spans .40--.90, so its visual midpoint is about .65. The caption now says so. \\
\addlinespace
``Confidence reliability'', ``monitoring quality'' used for Type-2 AUROC. & ``Confidence discrimination''. Type-2 AUROC measures discrimination, not calibration or reliability in the psychometric sense. \\
\addlinespace
Bootstrap intervals drawn from one random stream across all cells (seed 42). & Each cell seeded separately, so an interval depends only on its own data. Interval endpoints shift by up to a few hundredths. \\
\bottomrule
\end{tabular}
\end{table}

\section{Elicitation prompt template}
\label{app:prompt}

\begin{verbatim}
Answer this multiple-choice question.

{question}

First, state your answer (A, B, C, or D).
Then, state your confidence that your answer is correct as a number from 0
(pure guess) to 100 (certain).

Reply in EXACTLY this format:
Answer: [letter]
Confidence: [number]
\end{verbatim}

Each (model, item) pair used an independent conversation context. Greedy decoding (temperature 0) throughout.

\section{Answer and confidence parsing}
\label{app:parsing}

Answers and confidences were extracted from each response by fixed rules, applied in order with the first match winning. The code is released (\texttt{notebooks/atlas\_benchmark.py}), and a verification script re-runs it over the raw platform outputs and reproduces all 47{,}151 records.

\textbf{Answer.} (1) \texttt{Answer:\textbackslash s*([A-Da-d])}; (2) otherwise the first standalone letter A--D; (3) otherwise the answer is recorded as X. Thirteen responses (0.03\%) were X: GPT-oss-120B 5, Gemini 3.1 Flash-Lite 2, Gemma 3 27B 2, Haiku 4.5 2, Gemma 3 12B 1, Qwen3 Next 80B Instruct 1. Version 1 scored them as incorrect; version 2 excludes them. The fallback rule fired on 42 responses, 28 of them from Gemini 2.5 Flash. A first standalone letter can occasionally be an unintended one (for example the article ``A''); spot checks found the intended answer.

\textbf{Confidence.} (1) \texttt{Confidence:\textbackslash s*(\textbackslash d+)}, clipped to 0--100; (2) otherwise the first standalone integer of up to three digits in 0--100; (3) otherwise the value 50 is imputed. Rule 3 fired on 985 responses: 980 from GPT-oss-120B (65\% of its responses) and five from four other models. The raw outputs show that none of these five models ever states a confidence of 50, so each 50 in their series is a parser default. All 985 are excluded in version 2 (some coincide with X answers). Per-model counts for every route are released (\texttt{data/parsing\_fallback\_counts.csv}).

\textbf{Reasoning traces.} For DeepSeek-R1, whose output includes its reasoning, rule 1 captures the first ``Confidence:'' in the response, which can fall inside the reasoning rather than in the final answer. On 82 items (5.5\%) the captured value differs from the final stated confidence (mean 82.0 versus 87.4). Using the final values changes R1's aggregate AUROC from .769 to .766. The released data keep the values as originally extracted, and the discrepant items are listed in \texttt{data/thinking\_conf\_deepseek-r1-0528.csv}.

\textbf{Models and runs.} Gemma 4 26B A4B appears in the benchmark's model list but its run never started (no logs or outputs), so no data exist for it. No model was excluded on the basis of its responses.

\section{Full 33-model AUROC matrix}
\label{app:matrix}

Table~\ref{tab:matrix} reports Type-2 AUROC for every model-domain cell. Figure~\ref{fig:heatmap} shows the same values.

\begin{table}[!htbp]
\centering
\caption{Type-2 AUROC per model-domain cell, all 33 models. $n$ = scored items after excluding unparsed answers and parser-imputed confidences (16 models have fewer than 1{,}500; GPT-oss-120B$^\dagger$ retains 518 after 982 exclusions and should not be interpreted at domain level). Agg = AUROC over all of a model's items. Models grouped by family, sorted by Agg within family.}
\label{tab:matrix}
\small
\begin{tabular}{lcccccccc}
\toprule
\textbf{Model} & \textbf{$n$} & \textbf{Applied} & \textbf{Factual} & \textbf{Human.} & \textbf{Social} & \textbf{Formal} & \textbf{Science} & \textbf{Agg} \\
\midrule
Opus 4.6 & 1500 & .847 & .818 & .704 & .873 & .743 & .758 & .806 \\
Opus 4.5 & 1500 & .821 & .795 & .678 & .867 & .787 & .765 & .798 \\
Sonnet 4.5 & 1500 & .849 & .709 & .715 & .801 & .752 & .758 & .795 \\
Opus 4.7 & 1499 & .852 & .726 & .692 & .825 & .780 & .816 & .792 \\
Sonnet 4.6 & 1500 & .887 & .780 & .760 & .694 & .788 & .782 & .777 \\
Sonnet 4 & 1500 & .828 & .683 & .756 & .794 & .735 & .786 & .773 \\
Haiku 4.5 & 1498 & .829 & .720 & .793 & .727 & .805 & .706 & .767 \\
Opus 4.1 & 1500 & .838 & .615 & .711 & .781 & .691 & .636 & .708 \\
\midrule
DeepSeek-R1 & 1500 & .843 & .682 & .785 & .814 & .745 & .694 & .769 \\
DeepSeek V3.2 & 1371 & .713 & .769 & .698 & .669 & .769 & .696 & .734 \\
DeepSeek V3.1 & 1380 & .756 & .690 & .717 & .636 & .797 & .626 & .716 \\
\midrule
Gemini 3.1 Pro & 1346 & .841 & .773 & .764 & .776 & .645 & .640 & .765 \\
Gemini 3 Flash & 1500 & .800 & .673 & .662 & .794 & .681 & .690 & .731 \\
Gemini 2.5 Flash & 1360 & .801 & .765 & .644 & .748 & .637 & .664 & .728 \\
Gemini 2.5 Pro & 981 & .889 & .619 & .623 & .754 & .573 & .485 & .712 \\
Gemini 3.1 FLite & 1365 & .823 & .752 & .651 & .624 & .566 & .549 & .655 \\
Gemini 2.0 FLite & 1500 & .688 & .692 & .662 & .555 & .667 & .621 & .639 \\
Gemini 2.0 Flash & 1500 & .758 & .732 & .663 & .503 & .572 & .606 & .623 \\
\midrule
Gemma 4 31B & 1500 & .869 & .663 & .737 & .806 & .812 & .710 & .771 \\
Gemma 3 27B & 1497 & .585 & .674 & .573 & .528 & .569 & .598 & .565 \\
Gemma 3 4B & 1445 & .536 & .564 & .506 & .489 & .541 & .566 & .528 \\
Gemma 3 12B & 1446 & .533 & .613 & .522 & .464 & .554 & .500 & .505 \\
Gemma 3 1B & 1499 & .495 & .501 & .500 & .490 & .503 & .502 & .498 \\
\midrule
GPT-oss-120B$^\dagger$ & 518 & .829 & .855 & .611 & .801 & .515 & .760 & .800 \\
GPT-oss-20B & 1500 & .769 & .766 & .803 & .827 & .758 & .709 & .793 \\
GPT-5.4 mini & 1356 & .742 & .797 & .723 & .604 & .572 & .725 & .651 \\
GPT-5.4 & 1500 & .818 & .778 & .640 & .747 & .539 & .649 & .649 \\
GPT-5.4 nano & 1500 & .633 & .748 & .726 & .486 & .548 & .615 & .582 \\
\midrule
Qwen Think & 1500 & .726 & .680 & .714 & .768 & .644 & .757 & .745 \\
Qwen Coder & 1500 & .650 & .716 & .723 & .616 & .576 & .695 & .671 \\
Qwen 80B Inst & 1498 & .648 & .667 & .716 & .691 & .585 & .584 & .637 \\
Qwen 235B & 1500 & .644 & .622 & .711 & .650 & .567 & .637 & .621 \\
\midrule
GLM-5 & 598 & .602 & .576 & .818 & .790 & .609 & .555 & .697 \\
\bottomrule
\end{tabular}
\end{table}

\section{GPT-oss-120B}
\label{app:gptoss}

GPT-oss-120B produced a parseable confidence on 520 of 1{,}500 items, and 518 remain after the two further unparsed answers are excluded. The imputation rate differs by domain, from 59\% (Natural Science) to 74\% (Factual Recall) ($\chi^2(5) = 18.67$, $p = .002$), so the retained items are not a random sample of the item pool. Accuracy is similar on imputed and retained items (.900 and .896), but whether the model states a confidence at all may depend on item properties that also bear on discrimination. Its aggregate AUROC (.800) and domain values (Table~\ref{tab:matrix}) are reported for completeness. With 65--102 items per domain and intervals as wide as [.128, .918] (Formal), its domain profile should not be interpreted. Excluding the model entirely leaves the domain ordering unchanged (Friedman $\chi^2(5) = 26.43$, $p < .0001$, $n = 32$).

\section{Robustness analyses}
\label{app:robust}

\textbf{Pairwise contrasts.} Table~\ref{tab:pairs} gives Wilcoxon signed-rank tests over the 33 models for all 15 domain pairs, Holm-corrected.

\begin{table}[!htbp]
\centering
\caption{Pairwise domain contrasts (Wilcoxon signed-rank over 33 models; Holm correction across 15 pairs). Diff = mean AUROC difference, first minus second.}
\label{tab:pairs}
\small
\begin{tabular}{lrrr}
\toprule
\textbf{Pair} & \textbf{Diff} & \textbf{$p$} & \textbf{$p_{\text{Holm}}$} \\
\midrule
Applied vs Science & .088 & $<.0001$ & $<.0001$ \\
Applied vs Formal & .094 & $<.0001$ & $<.0001$ \\
Applied vs Social & .053 & .0003 & .005 \\
Applied vs Humanities & .062 & .001 & .011 \\
Factual vs Science & .042 & .005 & .057 \\
Applied vs Factual & .046 & .014 & .144 \\
Factual vs Formal & .048 & .035 & .318 \\
Formal vs Social & $-$.041 & .044 & .355 \\
Science vs Social & $-$.035 & .060 & .420 \\
Formal vs Humanities & $-$.033 & .063 & .420 \\
Humanities vs Science & .026 & .135 & .674 \\
Factual vs Humanities & .016 & .280 & 1.000 \\
Humanities vs Social & $-$.009 & .597 & 1.000 \\
Factual vs Social & .007 & .846 & 1.000 \\
Formal vs Science & $-$.007 & .874 & 1.000 \\
\bottomrule
\end{tabular}
\end{table}

\textbf{Rank stability.} Items were resampled with replacement within every model-domain cell (500 resamples), and domain means were re-ranked each time. Applied ranked first in 100\% of resamples. The probability of a bottom-two rank was .932 for Formal, .924 for Science, .132 for Humanities, .008 for Social, .004 for Factual and 0 for Applied.

\textbf{Error counts.} AUROC in a cell with few errors is unstable. Cells have a median of 20 errors (minimum 2); 28 cells have fewer than 10 and 5 have fewer than 5. Per-cell counts are in Table~\ref{tab:s1}.

\textbf{Subject level.} For the 47 subjects with at least 30 errors pooled across models, the seven Applied subjects have a mean AUROC rank of 19.6, against 28.8 for 11 Formal subjects and 21.1 for nine Science subjects. Applied subjects do not rank significantly higher than Formal and Science subjects combined (one-sided Mann--Whitney $U = 87$, $p = .18$). The domain-level ordering is a property of the bins in aggregate, not a uniform property of their subjects.

\section{Reliability}
\label{app:splithalf}

\begin{figure}[!htbp]
\centering
\includegraphics[width=0.6\textwidth]{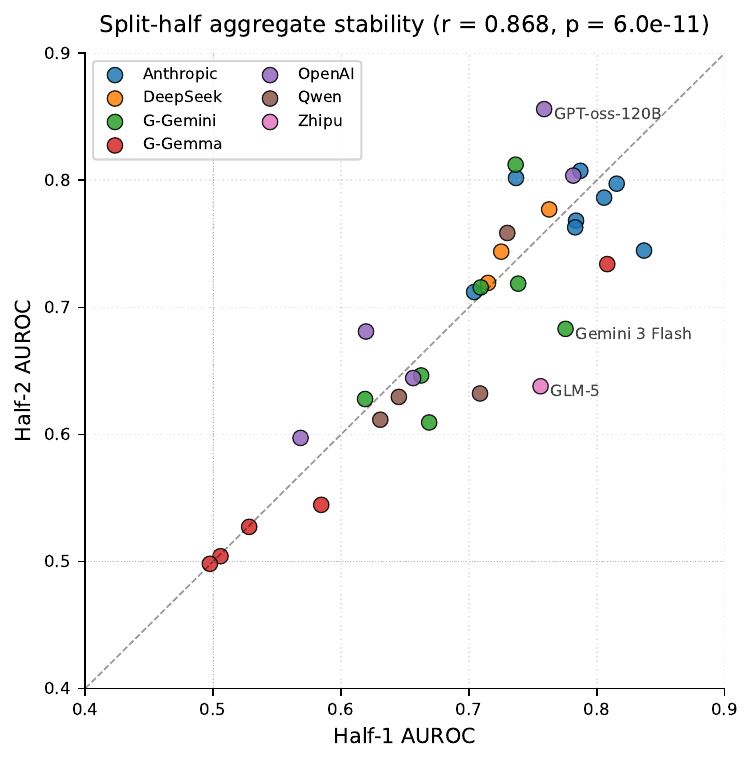}
\caption{Split-half stability of aggregate Type-2 AUROC across 33 models. Each point is one model; axes are AUROC on two random halves of its items. Cross-model $r = .868$.}
\label{fig:splithalf}
\end{figure}

Items were randomly split into halves, and AUROC was computed on each half for every model (Figure~\ref{fig:splithalf}). Aggregate AUROC correlates $r = .868$ across models between halves. For the population-level extremum contrast (Applied minus the mean of Formal and Science), the cross-model split-half correlation has median .249 over 50 splits (IQR .162--.343), and its population mean is positive in both halves in all 50. For full six-domain profiles, the within-model correlation between halves has median .157 over the same splits. Over 100 splits the grand median of per-model medians is .076, and 58\% of models are positive. These estimates move by several hundredths when a handful of records change, which is itself a sign of their imprecision.

\section{Families and generations}
\label{app:families}

\begin{figure}[!htbp]
\centering
\includegraphics[width=0.7\textwidth]{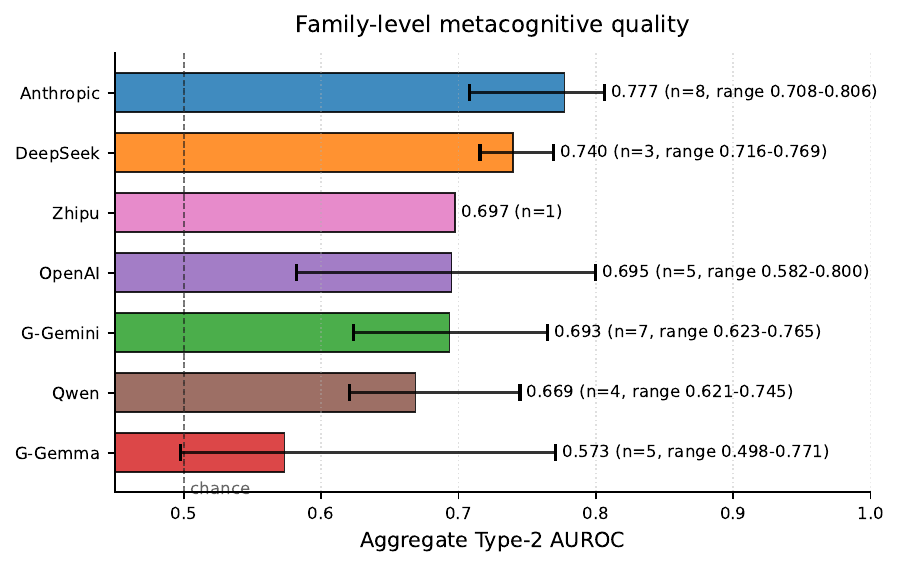}
\caption{Family mean domain profiles (Type-2 AUROC). Exploratory; see \S\ref{sec:hierarchy} on profile reliability.}
\label{fig:families}
\end{figure}

\textbf{Item-level transitions.} Table~\ref{tab:transitions} cross-classifies shared items by correctness under the earlier and later model in three generational pairs. Per-item confidence correlates more weakly across the Gemma step (Spearman $\rho = .36$) than across Anthropic steps (.75--.85) or DeepSeek steps (.66--.68).

\begin{table}[!htbp]
\centering
\caption{Correctness transitions on shared items (W = wrong, C = correct; earlier $\to$ later). The last column is the later model's mean confidence on items it newly gets wrong.}
\label{tab:transitions}
\small
\begin{tabular}{lrrrrr}
\toprule
\textbf{Pair} & \textbf{W$\to$W} & \textbf{W$\to$C} & \textbf{C$\to$W} & \textbf{C$\to$C} & \textbf{Conf.\ on C$\to$W} \\
\midrule
Gemma 3 27B $\to$ Gemma 4 31B & 57 & 265 & 32 & 1{,}144 & 90.8 \\
Opus 4.1 $\to$ Opus 4.7 & 49 & 37 & 31 & 1{,}382 & 71.5 \\
DeepSeek V3.2 $\to$ R1 & 71 & 64 & 48 & 1{,}188 & 84.6 \\
\bottomrule
\end{tabular}
\end{table}

\section{GLM-5 retest}
\label{app:glm5}

GLM-5's original run stopped after 598 items. It was rerun on the full item set three months later under the same prompt and settings. Table~\ref{tab:glm5} compares the two runs on the 598 shared items. The six domain values correlate $\rho = .77$ ($p = .072$) and $r = .75$. Correctness agrees on 98.5\% of items, and item-level confidence correlates $r = .71$. In both runs Humanities exceeds Applied. With 2 to 14 errors per domain, neither run's domain estimates are precise.

\begin{table}[!htbp]
\centering
\caption{GLM-5 on the same 598 items in two runs three months apart: errors and Type-2 AUROC per domain.}
\label{tab:glm5}
\small
\begin{tabular}{lrrrrr}
\toprule
& & \multicolumn{2}{c}{\textbf{Original run}} & \multicolumn{2}{c}{\textbf{Rerun}} \\
\cmidrule(lr){3-4}\cmidrule(lr){5-6}
\textbf{Domain} & \textbf{$n$} & \textbf{Errors} & \textbf{AUROC} & \textbf{Errors} & \textbf{AUROC} \\
\midrule
Applied & 84 & 4 & .602 & 5 & .639 \\
Factual & 111 & 8 & .576 & 10 & .702 \\
Formal & 102 & 3 & .609 & 5 & .757 \\
Humanities & 91 & 3 & .818 & 2 & .792 \\
Science & 110 & 3 & .555 & 4 & .709 \\
Social & 100 & 14 & .790 & 12 & .777 \\
\midrule
Aggregate & 598 & 35 & .697 & 38 & .759 \\
\bottomrule
\end{tabular}
\end{table}

\section{Partial runs}
\label{app:partial}

Twelve runs are incomplete because of platform timeouts; the smallest are GLM-5 (598 items) and Gemini 2.5 Pro (981). Four further models fall below 1{,}500 only because of the exclusions in Appendix~\ref{app:parsing}. Whether an item is present was tested against its domain for each incomplete model ($\chi^2$, 5 df). Presence is unrelated to domain ($p > .07$) for every model except Gemma 3 12B ($p = .010$; 54 items missing) and GPT-oss-120B ($p = .003$; Appendix~\ref{app:gptoss}).

\section{Future directions}

\textbf{Transport.} Whether the population-level ordering holds on a benchmark with non-overlapping items (for example GPQA) and under other elicitation formats is the main open question.

\textbf{Mechanism.} Hidden-state analyses of open-weight models (Gemma 4 31B, GPT-oss-20B, Qwen Think) could test whether Applied items produce more separable internal states than Formal items.

\textbf{Precision.} Per-model profiles would need many more errors per cell to be interpretable. For high-accuracy models that means many more items per domain, or harder items.

\textbf{Repair.} Whether targeted fine-tuning with confidence in the training signal can raise discrimination in a model's weakest domain is untested. The Gemma step shows that a large aggregate gain can come from error removal alone (Table~\ref{tab:generational}), so any such test should decompose its effect the same way.

\section{Procedural analogy to clinical assessment}
\label{app:neuropsych}

Clinical neuropsychological assessment follows a fixed interpretive sequence. Validity indicators are checked first \citep{larrabee2012}. If the protocol is valid, aggregate scores give an overall level, and the profile of index scores is then examined for relative strengths and weaknesses. The atlas follows the same order: screen the confidence signal \citep{cacioli2026e}, compute the aggregate, then examine the domain profile. The analogy is procedural only. Human cognitive domains rest on extensive factor-analytic validation; the MMLU bins used here do not, and \S\ref{sec:hierarchy} shows they are not cohesive. The analogy also marks a limit: in clinical practice an index score is interpreted only when its reliability supports it, and at current item counts most per-model domain profiles here do not meet that standard.

\section{Bootstrap intervals and error counts}
\label{app:ci}

Table~\ref{tab:s1} gives bootstrap 95\% intervals and error counts for all 198 model-domain cells. Median interval width is .202.

\begin{longtable}{llrrrrr}
\caption{Bootstrap 95\% confidence intervals and error counts for all 198 model-domain AUROC cells (1{,}000 resamples per cell, seeded per cell). Median CI width = .202. Archived as \texttt{data/atlas\_bootstrap\_cis.csv} and \texttt{data/atlas\_cell\_error\_counts.csv}.} \label{tab:s1} \\
\toprule
\textbf{Model} & \textbf{Domain} & \textbf{n} & \textbf{Errors} & \textbf{AUROC} & \textbf{CI lo} & \textbf{CI hi} \\
\midrule
\endfirsthead
\multicolumn{7}{c}{\tablename\ \thetable\ -- continued} \\
\toprule
\textbf{Model} & \textbf{Domain} & \textbf{n} & \textbf{Errors} & \textbf{AUROC} & \textbf{CI lo} & \textbf{CI hi} \\
\midrule
\endhead
\midrule
\multicolumn{7}{r}{{Continued on next page}} \\
\endfoot
\bottomrule
\endlastfoot
Opus 4.6 & Applied & 250 & 18 & 0.847 & 0.765 & 0.923 \\
Opus 4.6 & Factual & 250 & 15 & 0.818 & 0.694 & 0.927 \\
Opus 4.6 & Human. & 250 & 8 & 0.704 & 0.421 & 0.940 \\
Opus 4.6 & Social & 250 & 12 & 0.873 & 0.800 & 0.938 \\
Opus 4.6 & Formal & 250 & 8 & 0.743 & 0.446 & 0.977 \\
Opus 4.6 & Science & 250 & 13 & 0.758 & 0.623 & 0.886 \\
Opus 4.5 & Applied & 250 & 18 & 0.821 & 0.736 & 0.893 \\
Opus 4.5 & Factual & 250 & 16 & 0.795 & 0.657 & 0.907 \\
Opus 4.5 & Human. & 250 & 8 & 0.678 & 0.409 & 0.919 \\
Opus 4.5 & Social & 250 & 14 & 0.867 & 0.759 & 0.946 \\
Opus 4.5 & Formal & 250 & 10 & 0.787 & 0.565 & 0.950 \\
Opus 4.5 & Science & 250 & 13 & 0.764 & 0.629 & 0.884 \\
Sonnet 4.5 & Applied & 250 & 26 & 0.849 & 0.781 & 0.910 \\
Sonnet 4.5 & Factual & 250 & 13 & 0.709 & 0.528 & 0.857 \\
Sonnet 4.5 & Human. & 250 & 7 & 0.715 & 0.464 & 0.915 \\
Sonnet 4.5 & Social & 250 & 16 & 0.801 & 0.708 & 0.889 \\
Sonnet 4.5 & Formal & 250 & 7 & 0.752 & 0.507 & 0.989 \\
Sonnet 4.5 & Science & 250 & 12 & 0.758 & 0.593 & 0.925 \\
Opus 4.7 & Applied & 250 & 19 & 0.853 & 0.764 & 0.917 \\
Opus 4.7 & Factual & 249 & 14 & 0.726 & 0.581 & 0.868 \\
Opus 4.7 & Human. & 250 & 7 & 0.692 & 0.389 & 0.970 \\
Opus 4.7 & Social & 250 & 17 & 0.825 & 0.735 & 0.895 \\
Opus 4.7 & Formal & 250 & 12 & 0.780 & 0.592 & 0.943 \\
Opus 4.7 & Science & 250 & 11 & 0.817 & 0.677 & 0.934 \\
Sonnet 4.6 & Applied & 250 & 20 & 0.887 & 0.822 & 0.940 \\
Sonnet 4.6 & Factual & 250 & 17 & 0.780 & 0.646 & 0.903 \\
Sonnet 4.6 & Human. & 250 & 13 & 0.760 & 0.578 & 0.916 \\
Sonnet 4.6 & Social & 250 & 27 & 0.694 & 0.598 & 0.784 \\
Sonnet 4.6 & Formal & 250 & 8 & 0.788 & 0.540 & 0.976 \\
Sonnet 4.6 & Science & 250 & 12 & 0.782 & 0.629 & 0.906 \\
Sonnet 4 & Applied & 250 & 21 & 0.828 & 0.766 & 0.888 \\
Sonnet 4 & Factual & 250 & 16 & 0.683 & 0.543 & 0.815 \\
Sonnet 4 & Human. & 250 & 13 & 0.756 & 0.616 & 0.873 \\
Sonnet 4 & Social & 250 & 21 & 0.794 & 0.698 & 0.866 \\
Sonnet 4 & Formal & 250 & 9 & 0.735 & 0.558 & 0.915 \\
Sonnet 4 & Science & 250 & 16 & 0.786 & 0.665 & 0.893 \\
Haiku 4.5 & Applied & 248 & 28 & 0.829 & 0.743 & 0.907 \\
Haiku 4.5 & Factual & 250 & 26 & 0.720 & 0.610 & 0.819 \\
Haiku 4.5 & Human. & 250 & 11 & 0.793 & 0.645 & 0.915 \\
Haiku 4.5 & Social & 250 & 28 & 0.728 & 0.638 & 0.810 \\
Haiku 4.5 & Formal & 250 & 17 & 0.805 & 0.663 & 0.923 \\
Haiku 4.5 & Science & 250 & 16 & 0.706 & 0.567 & 0.838 \\
Opus 4.1 & Applied & 250 & 18 & 0.838 & 0.781 & 0.889 \\
Opus 4.1 & Factual & 250 & 11 & 0.615 & 0.444 & 0.775 \\
Opus 4.1 & Human. & 250 & 7 & 0.711 & 0.479 & 0.924 \\
Opus 4.1 & Social & 250 & 20 & 0.781 & 0.682 & 0.867 \\
Opus 4.1 & Formal & 250 & 12 & 0.691 & 0.521 & 0.840 \\
Opus 4.1 & Science & 250 & 18 & 0.636 & 0.526 & 0.761 \\
DeepSeek-R1 & Applied & 250 & 27 & 0.843 & 0.775 & 0.907 \\
DeepSeek-R1 & Factual & 250 & 20 & 0.682 & 0.539 & 0.815 \\
DeepSeek-R1 & Human. & 250 & 14 & 0.785 & 0.635 & 0.922 \\
DeepSeek-R1 & Social & 250 & 33 & 0.814 & 0.725 & 0.889 \\
DeepSeek-R1 & Formal & 250 & 13 & 0.745 & 0.596 & 0.875 \\
DeepSeek-R1 & Science & 250 & 20 & 0.694 & 0.540 & 0.830 \\
DeepSeek V3.2 & Applied & 228 & 31 & 0.713 & 0.612 & 0.803 \\
DeepSeek V3.2 & Factual & 221 & 23 & 0.769 & 0.683 & 0.849 \\
DeepSeek V3.2 & Human. & 229 & 13 & 0.698 & 0.562 & 0.817 \\
DeepSeek V3.2 & Social & 231 & 36 & 0.669 & 0.576 & 0.751 \\
DeepSeek V3.2 & Formal & 231 & 13 & 0.769 & 0.607 & 0.901 \\
DeepSeek V3.2 & Science & 231 & 19 & 0.697 & 0.557 & 0.811 \\
DeepSeek V3.1 & Applied & 234 & 44 & 0.756 & 0.681 & 0.820 \\
DeepSeek V3.1 & Factual & 229 & 22 & 0.690 & 0.578 & 0.799 \\
DeepSeek V3.1 & Human. & 230 & 20 & 0.717 & 0.591 & 0.825 \\
DeepSeek V3.1 & Social & 225 & 38 & 0.636 & 0.531 & 0.728 \\
DeepSeek V3.1 & Formal & 228 & 24 & 0.797 & 0.691 & 0.888 \\
DeepSeek V3.1 & Science & 234 & 29 & 0.626 & 0.514 & 0.733 \\
Gemini 3.1 Pro & Applied & 227 & 13 & 0.841 & 0.715 & 0.949 \\
Gemini 3.1 Pro & Factual & 215 & 14 & 0.773 & 0.629 & 0.899 \\
Gemini 3.1 Pro & Human. & 228 & 8 & 0.764 & 0.533 & 0.972 \\
Gemini 3.1 Pro & Social & 225 & 19 & 0.776 & 0.663 & 0.883 \\
Gemini 3.1 Pro & Formal & 230 & 6 & 0.645 & 0.469 & 0.891 \\
Gemini 3.1 Pro & Science & 221 & 9 & 0.640 & 0.467 & 0.826 \\
Gemini 3 Flash & Applied & 250 & 13 & 0.800 & 0.718 & 0.881 \\
Gemini 3 Flash & Factual & 250 & 13 & 0.673 & 0.516 & 0.825 \\
Gemini 3 Flash & Human. & 250 & 9 & 0.662 & 0.496 & 0.829 \\
Gemini 3 Flash & Social & 250 & 17 & 0.794 & 0.693 & 0.879 \\
Gemini 3 Flash & Formal & 250 & 9 & 0.681 & 0.504 & 0.862 \\
Gemini 3 Flash & Science & 250 & 12 & 0.690 & 0.530 & 0.845 \\
Gemini 2.5 Flash & Applied & 227 & 28 & 0.801 & 0.709 & 0.880 \\
Gemini 2.5 Flash & Factual & 219 & 20 & 0.765 & 0.638 & 0.883 \\
Gemini 2.5 Flash & Human. & 231 & 12 & 0.644 & 0.485 & 0.807 \\
Gemini 2.5 Flash & Social & 228 & 32 & 0.748 & 0.661 & 0.835 \\
Gemini 2.5 Flash & Formal & 230 & 19 & 0.637 & 0.534 & 0.745 \\
Gemini 2.5 Flash & Science & 225 & 16 & 0.664 & 0.546 & 0.800 \\
Gemini 2.5 Pro & Applied & 165 & 17 & 0.889 & 0.794 & 0.960 \\
Gemini 2.5 Pro & Factual & 172 & 11 & 0.619 & 0.484 & 0.788 \\
Gemini 2.5 Pro & Human. & 160 & 6 & 0.623 & 0.449 & 0.844 \\
Gemini 2.5 Pro & Social & 150 & 15 & 0.754 & 0.623 & 0.893 \\
Gemini 2.5 Pro & Formal & 163 & 6 & 0.573 & 0.478 & 0.777 \\
Gemini 2.5 Pro & Science & 171 & 7 & 0.485 & 0.469 & 0.497 \\
Gemini 3.1 FLite & Applied & 231 & 28 & 0.823 & 0.737 & 0.898 \\
Gemini 3.1 FLite & Factual & 233 & 18 & 0.752 & 0.636 & 0.858 \\
Gemini 3.1 FLite & Human. & 222 & 9 & 0.651 & 0.471 & 0.841 \\
Gemini 3.1 FLite & Social & 229 & 29 & 0.624 & 0.531 & 0.714 \\
Gemini 3.1 FLite & Formal & 227 & 31 & 0.566 & 0.506 & 0.641 \\
Gemini 3.1 FLite & Science & 223 & 19 & 0.549 & 0.474 & 0.646 \\
Gemini 2.0 FLite & Applied & 250 & 56 & 0.688 & 0.623 & 0.749 \\
Gemini 2.0 FLite & Factual & 250 & 35 & 0.692 & 0.585 & 0.777 \\
Gemini 2.0 FLite & Human. & 250 & 25 & 0.662 & 0.547 & 0.769 \\
Gemini 2.0 FLite & Social & 250 & 48 & 0.555 & 0.468 & 0.646 \\
Gemini 2.0 FLite & Formal & 250 & 61 & 0.667 & 0.597 & 0.740 \\
Gemini 2.0 FLite & Science & 250 & 45 & 0.621 & 0.533 & 0.711 \\
Gemini 2.0 Flash & Applied & 250 & 45 & 0.758 & 0.691 & 0.822 \\
Gemini 2.0 Flash & Factual & 250 & 29 & 0.732 & 0.629 & 0.822 \\
Gemini 2.0 Flash & Human. & 250 & 13 & 0.663 & 0.512 & 0.808 \\
Gemini 2.0 Flash & Social & 250 & 41 & 0.503 & 0.420 & 0.583 \\
Gemini 2.0 Flash & Formal & 250 & 43 & 0.572 & 0.498 & 0.648 \\
Gemini 2.0 Flash & Science & 250 & 30 & 0.606 & 0.505 & 0.696 \\
Gemma 4 31B & Applied & 250 & 17 & 0.869 & 0.773 & 0.942 \\
Gemma 4 31B & Factual & 250 & 19 & 0.663 & 0.535 & 0.792 \\
Gemma 4 31B & Human. & 250 & 11 & 0.737 & 0.564 & 0.889 \\
Gemma 4 31B & Social & 250 & 23 & 0.806 & 0.703 & 0.904 \\
Gemma 4 31B & Formal & 250 & 9 & 0.812 & 0.638 & 0.981 \\
Gemma 4 31B & Science & 250 & 12 & 0.710 & 0.558 & 0.856 \\
Gemma 3 27B & Applied & 248 & 67 & 0.585 & 0.526 & 0.646 \\
Gemma 3 27B & Factual & 250 & 37 & 0.674 & 0.599 & 0.748 \\
Gemma 3 27B & Human. & 249 & 24 & 0.573 & 0.488 & 0.665 \\
Gemma 3 27B & Social & 250 & 54 & 0.528 & 0.468 & 0.583 \\
Gemma 3 27B & Formal & 250 & 76 & 0.569 & 0.500 & 0.634 \\
Gemma 3 27B & Science & 250 & 63 & 0.598 & 0.543 & 0.645 \\
Gemma 3 4B & Applied & 244 & 118 & 0.536 & 0.485 & 0.585 \\
Gemma 3 4B & Factual & 241 & 65 & 0.564 & 0.494 & 0.637 \\
Gemma 3 4B & Human. & 240 & 55 & 0.506 & 0.441 & 0.562 \\
Gemma 3 4B & Social & 240 & 113 & 0.488 & 0.430 & 0.543 \\
Gemma 3 4B & Formal & 239 & 131 & 0.541 & 0.483 & 0.609 \\
Gemma 3 4B & Science & 241 & 112 & 0.566 & 0.507 & 0.621 \\
Gemma 3 12B & Applied & 242 & 77 & 0.533 & 0.463 & 0.601 \\
Gemma 3 12B & Factual & 237 & 47 & 0.614 & 0.522 & 0.696 \\
Gemma 3 12B & Human. & 243 & 29 & 0.522 & 0.420 & 0.625 \\
Gemma 3 12B & Social & 247 & 71 & 0.464 & 0.392 & 0.537 \\
Gemma 3 12B & Formal & 233 & 99 & 0.554 & 0.502 & 0.605 \\
Gemma 3 12B & Science & 244 & 66 & 0.500 & 0.430 & 0.575 \\
Gemma 3 1B & Applied & 249 & 163 & 0.495 & 0.465 & 0.524 \\
Gemma 3 1B & Factual & 250 & 124 & 0.501 & 0.472 & 0.531 \\
Gemma 3 1B & Human. & 250 & 112 & 0.500 & 0.476 & 0.523 \\
Gemma 3 1B & Social & 250 & 161 & 0.490 & 0.463 & 0.514 \\
Gemma 3 1B & Formal & 250 & 185 & 0.503 & 0.477 & 0.527 \\
Gemma 3 1B & Science & 250 & 172 & 0.502 & 0.479 & 0.520 \\
GPT-oss-120B & Applied & 93 & 18 & 0.829 & 0.742 & 0.902 \\
GPT-oss-120B & Factual & 65 & 7 & 0.855 & 0.625 & 0.989 \\
GPT-oss-120B & Human. & 81 & 5 & 0.611 & 0.244 & 0.925 \\
GPT-oss-120B & Social & 78 & 11 & 0.801 & 0.659 & 0.923 \\
GPT-oss-120B & Formal & 99 & 2 & 0.515 & 0.128 & 0.918 \\
GPT-oss-120B & Science & 102 & 11 & 0.760 & 0.582 & 0.900 \\
GPT-oss-20B & Applied & 250 & 57 & 0.769 & 0.697 & 0.827 \\
GPT-oss-20B & Factual & 250 & 34 & 0.766 & 0.675 & 0.847 \\
GPT-oss-20B & Human. & 250 & 25 & 0.803 & 0.688 & 0.904 \\
GPT-oss-20B & Social & 250 & 27 & 0.827 & 0.745 & 0.901 \\
GPT-oss-20B & Formal & 250 & 16 & 0.758 & 0.588 & 0.896 \\
GPT-oss-20B & Science & 250 & 20 & 0.709 & 0.592 & 0.810 \\
GPT-5.4 mini & Applied & 227 & 45 & 0.743 & 0.649 & 0.821 \\
GPT-5.4 mini & Factual & 216 & 26 & 0.797 & 0.712 & 0.870 \\
GPT-5.4 mini & Human. & 230 & 14 & 0.723 & 0.598 & 0.834 \\
GPT-5.4 mini & Social & 228 & 55 & 0.604 & 0.508 & 0.693 \\
GPT-5.4 mini & Formal & 232 & 64 & 0.572 & 0.492 & 0.649 \\
GPT-5.4 mini & Science & 223 & 41 & 0.725 & 0.631 & 0.815 \\
GPT-5.4 & Applied & 250 & 29 & 0.818 & 0.735 & 0.887 \\
GPT-5.4 & Factual & 250 & 21 & 0.777 & 0.652 & 0.880 \\
GPT-5.4 & Human. & 250 & 11 & 0.640 & 0.397 & 0.876 \\
GPT-5.4 & Social & 250 & 27 & 0.747 & 0.638 & 0.849 \\
GPT-5.4 & Formal & 250 & 41 & 0.539 & 0.452 & 0.631 \\
GPT-5.4 & Science & 250 & 28 & 0.649 & 0.533 & 0.754 \\
GPT-5.4 nano & Applied & 250 & 85 & 0.633 & 0.565 & 0.700 \\
GPT-5.4 nano & Factual & 250 & 58 & 0.749 & 0.674 & 0.813 \\
GPT-5.4 nano & Human. & 250 & 50 & 0.726 & 0.652 & 0.798 \\
GPT-5.4 nano & Social & 250 & 100 & 0.486 & 0.410 & 0.564 \\
GPT-5.4 nano & Formal & 250 & 132 & 0.548 & 0.477 & 0.620 \\
GPT-5.4 nano & Science & 250 & 95 & 0.615 & 0.538 & 0.691 \\
Qwen Think & Applied & 250 & 22 & 0.726 & 0.611 & 0.838 \\
Qwen Think & Factual & 250 & 18 & 0.680 & 0.528 & 0.811 \\
Qwen Think & Human. & 250 & 10 & 0.714 & 0.529 & 0.875 \\
Qwen Think & Social & 250 & 32 & 0.768 & 0.695 & 0.836 \\
Qwen Think & Formal & 250 & 8 & 0.644 & 0.443 & 0.838 \\
Qwen Think & Science & 250 & 14 & 0.757 & 0.630 & 0.877 \\
Qwen Coder & Applied & 250 & 49 & 0.650 & 0.585 & 0.716 \\
Qwen Coder & Factual & 250 & 29 & 0.716 & 0.637 & 0.791 \\
Qwen Coder & Human. & 250 & 16 & 0.723 & 0.597 & 0.827 \\
Qwen Coder & Social & 250 & 51 & 0.616 & 0.540 & 0.688 \\
Qwen Coder & Formal & 250 & 23 & 0.576 & 0.459 & 0.690 \\
Qwen Coder & Science & 250 & 28 & 0.696 & 0.603 & 0.794 \\
Qwen 80B Inst & Applied & 250 & 45 & 0.647 & 0.571 & 0.723 \\
Qwen 80B Inst & Factual & 249 & 28 & 0.667 & 0.571 & 0.770 \\
Qwen 80B Inst & Human. & 250 & 12 & 0.716 & 0.564 & 0.854 \\
Qwen 80B Inst & Social & 250 & 49 & 0.691 & 0.621 & 0.759 \\
Qwen 80B Inst & Formal & 250 & 46 & 0.585 & 0.516 & 0.671 \\
Qwen 80B Inst & Science & 249 & 35 & 0.584 & 0.483 & 0.680 \\
Qwen 235B & Applied & 250 & 55 & 0.644 & 0.581 & 0.711 \\
Qwen 235B & Factual & 250 & 23 & 0.622 & 0.522 & 0.728 \\
Qwen 235B & Human. & 250 & 16 & 0.711 & 0.601 & 0.827 \\
Qwen 235B & Social & 250 & 33 & 0.650 & 0.538 & 0.740 \\
Qwen 235B & Formal & 250 & 42 & 0.567 & 0.478 & 0.655 \\
Qwen 235B & Science & 250 & 32 & 0.637 & 0.573 & 0.695 \\
GLM-5 & Applied & 84 & 4 & 0.602 & 0.221 & 0.916 \\
GLM-5 & Factual & 111 & 8 & 0.576 & 0.399 & 0.775 \\
GLM-5 & Human. & 91 & 3 & 0.818 & 0.706 & 0.972 \\
GLM-5 & Social & 100 & 14 & 0.790 & 0.648 & 0.908 \\
GLM-5 & Formal & 102 & 3 & 0.609 & 0.388 & 1.000 \\
GLM-5 & Science & 110 & 3 & 0.554 & 0.387 & 0.853 \\
\end{longtable}

\end{document}